\documentclass[sigconf]{acmart}
\AtBeginDocument{%
  \providecommand\BibTeX{{%
    \normalfont B\kern-0.5em{\scshape i\kern-0.25em b}\kern-0.8em\TeX}}}
\newcommand{\ie}{\textit{i}.\textit{e}.}
\newcommand{\eg}{\textit{e}.\textit{g}.}

\usepackage{pifont}
\usepackage{arydshln}

\newcommand{\xmark}{\ding{55}} 
\newcommand{\myparagraph}[1]{\vspace{0.1em}\noindent\textbf{#1}}

\setcopyright{acmcopyright}
\copyrightyear{2022}
\acmYear{2022}
\acmDOI{10.1145/3503161.3547858} 
\acmConference[MM'22]{Make sure to enter the correct conference title from your rights confirmation emai}{October, 10--14, 2022}{Lisbon, Portugal}
\acmBooktitle{Proceedings of the 30th ACM International Conference on Multimedia (MM'22), October, 10--14, 2022. Lisbon, Portugal} 
\acmPrice{15.00}
\acmISBN{978-1-4503-9203-7/22/10}
\begin{document}
\title{Graph Reasoning Transformer for Image Parsing}
\author{Dong Zhang}
\affiliation{%
\institution{The Hong Kong University of Science and Technology}
\city{Hong Kong}
\country{China}}
\email{dongz@ust.hk}
\author{Jinhui Tang}
\affiliation{%
\institution{Nanjing University of Science and Technology}
\city{Nanjing}
\country{China}}
\email{jinhuitang@njust.edu.cn}
\author{Kwang-Ting Cheng}
\affiliation{%
\institution{The Hong Kong University of Science and Technology}
\city{Hong Kong}
\country{China}}
\email{timcheng@ust.hk}
\begin{abstract}
Capturing the long-range dependencies has empirically proven to be effective on a wide range of computer vision tasks. The progressive advances on this topic have been made through the employment of the transformer framework with the help of the multi-head attention mechanism. However, the attention-based image patch interaction potentially suffers from problems of \emph{redundant interactions of intra-class patches} and \emph{unoriented interactions of inter-class patches}. In this paper, we propose a novel Graph Reasoning Transformer (GReaT) for image parsing to enable image patches to interact following a relation reasoning pattern. Specifically, the linearly embedded image patches are first projected into the graph space, where each node represents the implicit visual center for a cluster of image patches and each edge reflects the relation weight between two adjacent nodes. After that, global relation reasoning is performed on this graph accordingly. Finally, all nodes including the relation information are mapped back into the original space for subsequent processes. Compared to the conventional transformer, GReaT has higher interaction efficiency and a more purposeful interaction pattern. Experiments are carried out on the challenging Cityscapes and ADE20K datasets. Results show that GReaT achieves consistent performance gains with slight computational overheads on the state-of-the-art transformer baselines.
\end{abstract}
\maketitle
\section{Introduction}\label{sec:1}
Image Parsing (IP) is a fundamental yet challenging research task in the community of multimedia and computer vision, which aims to assign each pixel of the input image a unique category label. In the past years, this task has been intensively studied and applied to various applications, \eg, autonomous driving~\cite{prakash2021multi}, computer-aided diagnosis~\cite{li2021superpixel}, and
makeup transfer~\cite{deng2021spatially}.

\begin{figure}
\vspace{6mm}
\includegraphics[width=.47\textwidth]{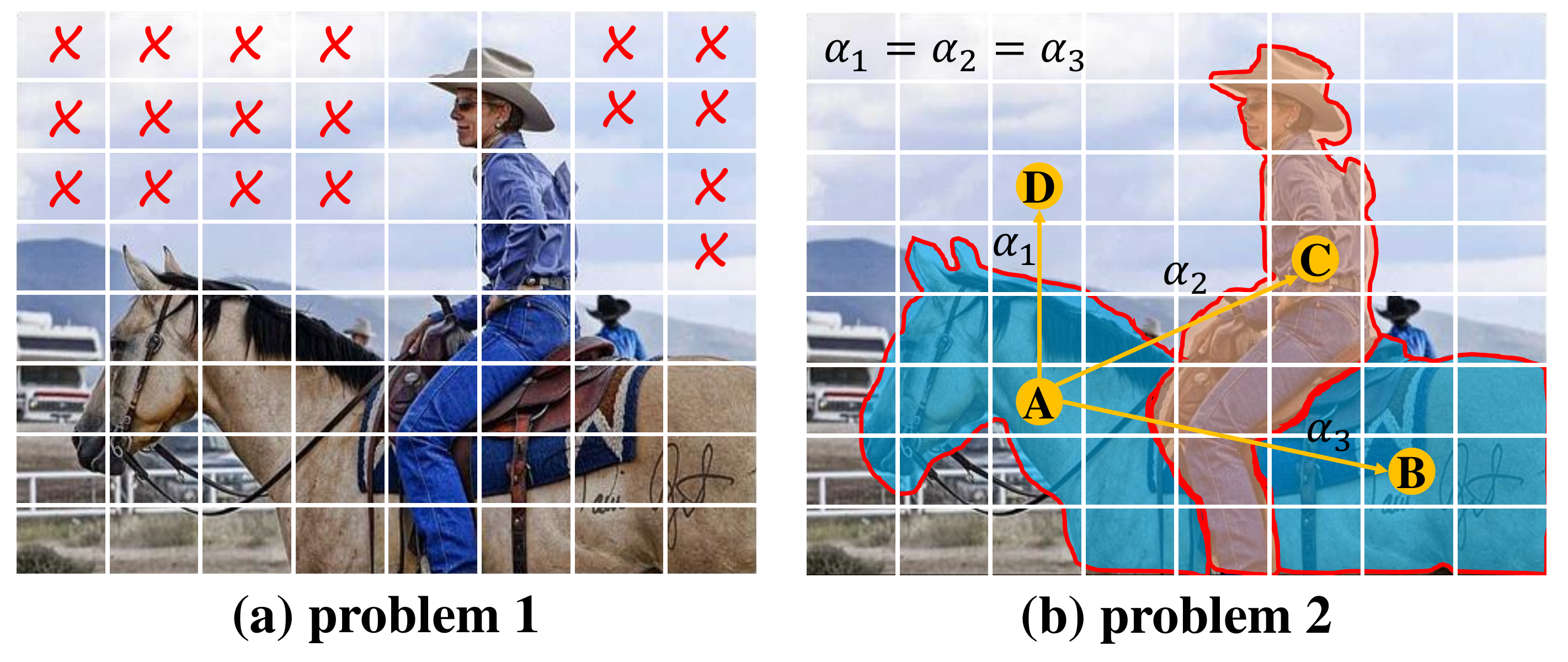}
\vspace{-2mm}
\caption{The key motivations of this work: there are two potential problems in the existing vision transformer. (a) \emph{redundant interactions of intra-class patches}. (b) \emph{unoriented interactions of inter-class patches}.}
\label{fig0}
\end{figure}

Thanks to the tremendous advances of Convolutional Neural Network (CNN) in image processing~\cite{lecun2015deep,he2016deep,tan2019efficientnet,xie2017aggregated}, successful IP models~\cite{badrinarayanan2017segnet,ronneberger2015u,tian2019decoders} are mainly built on Fully Convolutional Network (FCN)~\cite{long2015fully} with a CNN as the backbone. However, due to the limited local receptive fields of the standard convolution operations, FCN can only capture short-range dependencies (also known as the local contexts) of the given image, which are insufficient for some complex and diverse scenes. To alleviate this problem, a number of reformative methods~\cite{hou2020strip,zhao2017pyramid,chen2017deeplab,fu2019dual,zhang2021self,zhang2018context,huang2019ccnet,zhang2019co,peng2017large,yu2015multi} have been proposed. These methods stand on the shoulders of FCN with the objective of capturing the global long-range dependencies by either actively expanding the effective receptive fields~\cite{yu2015multi,peng2017large,zhang2020feature,zhang2021self} of the backbone or using some specific global context modeling schemes~\cite{hou2020strip,huang2019ccnet,zhao2017pyramid,chen2017deeplab,fu2019dual,zhang2018context,zhang2019co}, or both.

Despite the limited success of FCN and its extensions specifically targeting IP, the inherent locality problem in convolutions still exists.
Recently, inspired by the mature applications of the transformer framework~\cite{vaswani2017attention,devlin2018bert,zhang2020accelerating} on natural language processing, vision transformer has been studied extensively in the computer vision domain and has achieved a number of dazzling results on both images and videos~\cite{carion2020end,dosovitskiy2020image,liu2021swin,zheng2021rethinking,xie2021segformer,touvron2021training,wang2021pyramid}.
For a vision transformer model, it mainly consists of a patch partition operation, patch/position embedding layers, layer norms, multi-head attention layers, multi-layer perception layers, and some task-specific operations (\eg, vectorization of feature maps~\cite{dosovitskiy2020image}, multi-scale operation~\cite{wang2021pyramid} and patch merging operation~\cite{zheng2021rethinking}). As one of the core components, the multi-head attention is implemented in an unbiased, fully connected pattern for image patch interactions, which can capture the long-range dependencies (\ie, the global contexts) of the input. Therefore, the inherent locality problem in convolution operations can be completely solved in a vision transformer.

\begin{figure*}
\includegraphics[width=.90\textwidth]{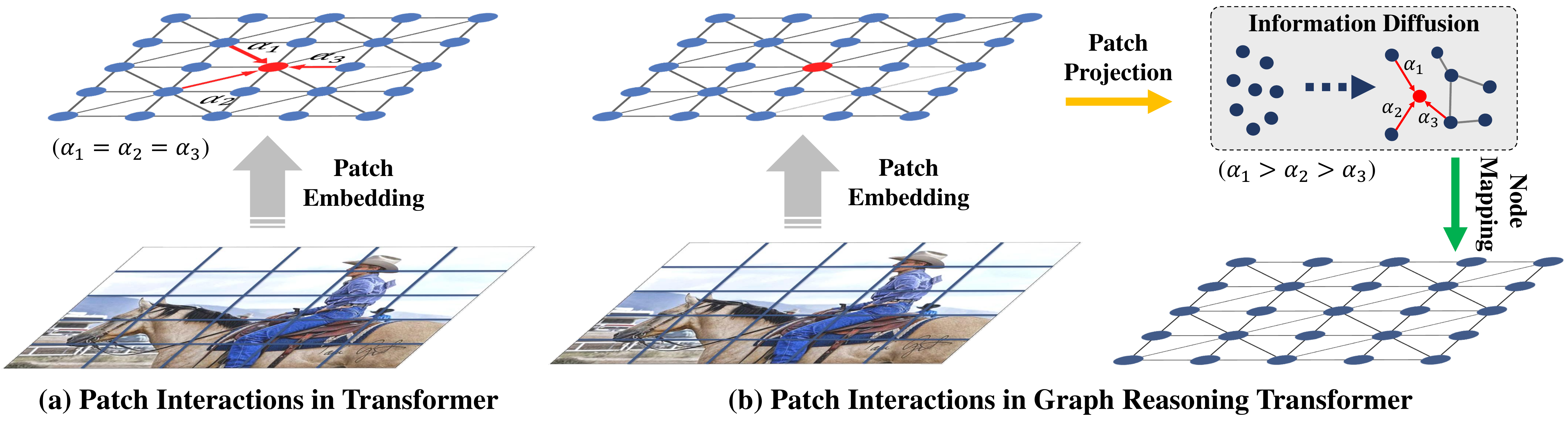}
\vspace{-3mm}
\caption{An illustration of our main idea. (a) Image patches in the conventional vision transformer interact following an unbiased fully connected pattern, which suffers from problems of \emph{redundant interactions of intra-class patches} and \emph{unoriented interactions of inter-class patches}. (b) GReaT enables image patches to interact following a global relation reasoning pattern. Compared to the conventional transformer, GReaT has higher interaction efficiency and a purposeful interaction pattern.}
\vspace{-2mm}
\label{fig1}
\end{figure*}

However, the existing patch interaction in the multi-head attention potentially suffers from the following two problems: 1) \emph{redundant interactions of intra-class patches} and 2) \emph{unoriented interactions of inter-class patches}. They are the key motivations of this paper.
For problem 1, as illustrated in Figure~\ref{fig0} (a), some image patches (marked with \textbf{\color{red}\xmark}) belonging to the same category (the ``sky'') do not contain any object boundary information, and thus the interaction among them would not be informative and necessary. This is also why spatial dropout/dropblock~\cite{choe2020attention,ghiasi2018dropblock} and token reorganization~\cite{liang2022not} are effective in vision recognition.
For problem 2, the existing patch interactions under the help of the multi-head attention mechanism do not distinguish among different object categories and are performed in a roughly unbiased manner. For example, as illustrated in Figure~\ref{fig0} (b), image patches B, C and D (which contain the ``horse'', the ``person'', and the ``sky'' categories respectively) are equal important to patch A (which contains the ``horse'' category) in the current interaction, \ie, $\alpha_1 = \alpha_2 = \alpha_3$ (where $\alpha_*$ denotes the interaction weight between two patches). However, this does not match the common sense~\cite{wang2020visual,zhang2020causal}. The interactions between one part of the``horse'' with another part, and between the ``horse'' and the ``person'' should be much more important than the interaction between the ``horse'' and the ``sky''.

To address the above two problems, we propose a novel Graph Reasoning Transformer (GReaT) model. Compared with the conventional transformer, GReaT has higher interaction efficiency and a more purposeful interaction pattern. Different from the existing patch interaction in the multi-head attention (as illustrated in Figure~\ref{fig1} (a)) module, we propose a Graph Reasoning Block (GReaB) for the vision transformer, which enables image patches to interact following a global relation reasoning pattern. Specifically, as illustrated in Figure~\ref{fig1} (b), the linearly embedded image patches are first projected into a graph representation via a patch projection operation, where each node represents the implicit visual center for a cluster of image patches and each edge reflects the relation weight between two adjacent nodes. It is then followed by the global relation reasoning on this graph by an information diffusion procedure. Finally, all nodes including the relation weight information are mapped back into the original space via the node mapping operation for subsequent processes. GReaT can be obtained by replacing the attention module in a vision transformer baseline model with the proposed GReaB. To demonstrate the superiority of our GReaT, experiments are carried out on the challenging Cityscapes~\cite{cordts2016cityscapes} and ADE20K~\cite{zhou2017scene} datasets. Results show that GReaT can bring achieve consistent performance gains with slight computational overheads on the state-of-the-art baseline models.

The main contributions of this paper are:
1) a novel interaction module for vision transformer to enable image patches to interact in the graph space, and 2) deploying a Graph Reasoning Block on a vision transformer baseline and achieving the competitive performance on two public IP datasets.

\section{Related Work}
\noindent
\myparagraph{Image Parsing (IP).} As one of the fundamental computer vision tasks (\eg, image classification~\cite{he2016deep}, object detection~\cite{ren2015faster}, object localization ~\cite{tompson2015efficient}, and instance segmentation~\cite{wang2021end} and IP~\cite{long2015fully}), IP has been intensively studied and made great advances in the past few years. Based on the idea of FCN~\cite{long2015fully}, the existing IP methods can be mainly divided into the following three camps: 1) CNN-based methods, 2) transformer-based methods and the hybrid (\ie, mixed CNN and transformer) methods. 
In the first camp, these methods mainly use a CNN as the backbone, and add some specific operations for upsampling~\cite{badrinarayanan2017segnet,yu2015multi,zhang2021self,ronneberger2015u,tian2019decoders,wang2020deep,lin2017refinenet} or context aggregation~\cite{hou2020strip,zhao2017pyramid,chen2017deeplab,fu2019dual,zhang2018context,huang2019ccnet,zhang2019co,peng2017large}. 
In the second camp, the input image is first divided into image patches, and then converted into sequences. On this basis, the transformer encoding is then completed via a series of repeated operations (\eg, layer norm, patch interaction and residual connection). Finally, upsampling and patch merging operations are deployed on the encoded image sequences before the model output. Methods~\cite{carion2020end,dosovitskiy2020image,liu2021swin,zheng2021rethinking,xie2021segformer,touvron2021training,wang2021pyramid} in this camp have the advantage of being inherently able to obtain long-range dependencies. In the third camp, methods are mainly based on utilizing advantages of both CNN and transformer at the same time as their starting point, \eg, TransUnet~\cite{chen2021transunet}, ConFormer~\cite{gulati2020conformer} and nnFormer~\cite{zhou2021nnformer}. In this work, we follow the transformer-based framework for IP. 

\noindent
\myparagraph{Vision Transformer.} 
Since ViT~\cite{dosovitskiy2020image} was successfully used in image classification, the transformer-based vision recognition models have been extended to a large number of computer vision tasks, \eg, object detection~\cite{carion2020end}, instance segmentation~\cite{wang2021end}, and object tracking~\cite{meinhardt2021trackformer}. For a computer vision transformer model, improving the computational efficiency of the multi-head attention module is one of the most key research topics. To this end, an intuitive approach is to shorten the image sequence length as in~\cite{wang2021pyramid,wang2018non}. However, this shoddy approach may lead to the problem that some critical feature cues are lost, which is particularly critical for the current IP models.
To retain as much features as possible while reducing the computational costs, some efficient attention methods are also proposed for vision transformer, \eg, dynamic token~\cite{wang2021not}, shifted windows~\cite{liu2021swin} and focal attention~\cite{yang2021focal}. Although the above methods can alleviate the low efficiency problem, the problem of the unoriented interactions of inter-class patches in the vision transformer still exists (cf. Figure~\ref{fig0} (b)). In this paper, we propose to use the global relation reasoning for patch interaction.

\noindent
\myparagraph{Long-Range Dependency.} 
In the era of deep learning, the previous methods mainly obtain long-range dependencies by increasing the effective receptive fields (\eg, dilated/atrous convolution~\cite{yu2015multi}, self-regulation~\cite{zhang2021self,TangLPT20} and large kernel operation~\cite{peng2017large}), or use the multi-scale features (PPM~\cite{zhao2017pyramid}, ASPP~\cite{chen2017deeplab} and MPM~\cite{hou2020strip}). In recent years, inspired by the non-local mean operation~\cite{buades2005non}, the progressive 
studies~\cite{zhang2019co,hou2020strip,fu2019dual,huang2019ccnet,dosovitskiy2020image,liu2021swin,Wang_2022_IJCAI,yan2018participation,zheng2021rethinking,xie2021segformer,yan2020higcin,touvron2021training,wang2021pyramid,TANG2022108792} mainly use a multi-head attention operation~\cite{vaswani2017attention,wang2018non} as the way to obtain the long-range dependencies. One of the core operations of the multi-head attention is to calculate similarities among image pixels, and redistribute similarities (\ie, weights) to each pixel to achieve the purpose of a global interaction. Although this mechanism can solve the inherent locality problem in CNN, it is virtually unreasonable to use it in a vision transformer for image patch interactions (as discussed in Section~\ref{sec:1}). In this paper, instead of using the conventional attention-based model, we propose to obtain long-range dependencies in the graph space.

\noindent
\myparagraph{Graph Reasoning (GR).} 
GR is one of the effective ways to capture the pixel-level long-range dependencies (\ie, the reciprocal co-occurrent features) of a given image. 
The GR methods can be divided into those without an external knowledge base~\cite{lin2020graphonomy,jain2021representing,plath2009multi,rong2020self,bertasius2017convolutional,wang2018non,chen2019graph,liang2018symbolic} and those with an external knowledge base~\cite{deng2014large,redmon2017yolo9000,sabour2017dynamic,zhao2017open}. In this paper, our method is also implemented belongs to the first type. In this domain, successful methods (\eg, conditional random field~\cite{plath2009multi} and random walk operation~\cite{bertasius2017convolutional}) have been used in IP and prediction masks with a satisfactory recognition performance, which are usually regarded as the post-processing steps in a fully-supervised model. Recently, the graph convolution operation (\eg, non-local~\cite{wang2018non}, GloRe unit~\cite{chen2019graph}, and SGR~\cite{liang2018symbolic}) using a structured densely connected graph (is also named as the affinity matrix) is proposed and swimmingly used in a number of computer vision tasks, \eg, classification~\cite{kipf2016semi,chen2019graph}, instance segmentation~\cite{li2018beyond,liang2018symbolic} and object detection~\cite{wang2018videos,yan2020social}. A common characteristic of these methods is that they are trained in an end-to-end manner and have the advantage of being plug-and-play. In this paper, our method is implicitly inspired by~\cite{chen2019graph,jain2021representing,liang2018symbolic,li2018beyond}, and our contribution lies in using GR to solve two potential problems in image patch interaction of the vision transformer. 
\section{Methodology}
In this section, we first make a general preliminary on the vision transformer framework for image parsing in Section~\ref{sec3:1}. We then introduce motivations and the overall architecture of our proposed Graph Reasoning Transformer (GReaT) in Section~\ref{sec3:2}. Finally, we show implementation details of the proposed Graph Reasoning Block (GReaB) in Section~\ref{sec3:3}.
\subsection{Preliminaries}\label{sec3:1}
Vision transformer is proposed to mainly make up shortcomings of the traditional CNN model in capturing long-range dependencies. We revisit the vision transformer for image parsing. For a given image $\textbf{I} \in \mathbb{R}^{H \times W \times C}$, we first use a patch partition operation to divide it into $N$ image patches, and of which is expressed as $\textbf{P}_n \in \mathbb{R}^{L \times L \times C}$, where $H$ and $W$ denotes the image height and width, respectively. $C$ denotes the channel size, and $L$ denotes the image patch resolution in both height and width. $n$ denotes the $n$-th image patch and $n=1,2,...,N$. Therefore, there are $N=HW/L^2$ patches, which are used as the input of the transformer model. After the patch is flattened into a 2D sequence and linearly embedded into the feature space (\ie, $\textbf{P}_n \rightarrow \textbf{X}_n^{\textrm{patch}} \in \mathbb{R}^{L^2 \times C'}$), we then add a learnable relative position encoding to each sequence to ensure that the spatial information of each patch can be preserved. This process can be formulated as:
\begin{equation}
\textbf{X}_n = \textrm{pos}(\textbf{X}_n^{\textrm{patch}}) + \textbf{X}_n^{\textrm{patch}},
\label{eq1}
\end{equation}
where $\textrm{pos}(\cdot)$ denotes the learnable relative position encoding layer. $\textbf{X}_n$ denotes an image patch including the relative position encoding information. For convenience, we omit patch flatten and linear embedding in Eq.~(\ref{eq1}). After that, the layer norm and the multi-head attention operations are performed on $\textbf{X}_n$ for normalization and interaction, respectively. With the help of a residual connection, we can obtain the interacted image patch:
\begin{equation}
\textbf{Y}_n = \textrm{MHA}(\textrm{Norm}(\textbf{X}_n)) + \textbf{X}_n,
\label{eq2}
\end{equation}
where $\textrm{Norm}(\cdot)$ denotes the layer norm operation, $\textrm{MHA}(\cdot)$ denotes the multi-head attention operation across all the image patches. $\textbf{Y}_n$ denotes the patch after interacting with other image patches. Then, a layer norm operation and a feed-forward network are implemented on $\textbf{Y}_n$. After a residual connection from $\textbf{Y}_n$, the current output can be obtained by:
\begin{equation}
\textbf{Z}_n = \textrm{FFN}(\textrm{Norm}(\textbf{Y}_n)) + \textbf{Y}_n,
\label{eq3}
\end{equation}
where $\textrm{FFN}(\cdot)$ denotes the feed-forward network, and $\textbf{Z}_n$ denotes the current output. In a vision transformer model, the above steps (\ie, from Eq.~(\ref{eq1}) to Eq.~(\ref{eq3})) are cascaded to form a holistic layer, termed as the transformer encoding. When we implement this transformer encoding layer multiple times (the residual connection and the downsampling operation are also contained if needed), a transformer encoder is formed, which can be used to extract the semantic patch features of the input image. Compared to a CNN backbone, patch features have more and abundant long-range dependency information. Finally, these features are reshaped and upsampled to the same spatial resolution as the input patch and used for predictions after a patch merging operation.
\begin{figure*}[t]
\centering
\includegraphics[width=.95\textwidth]{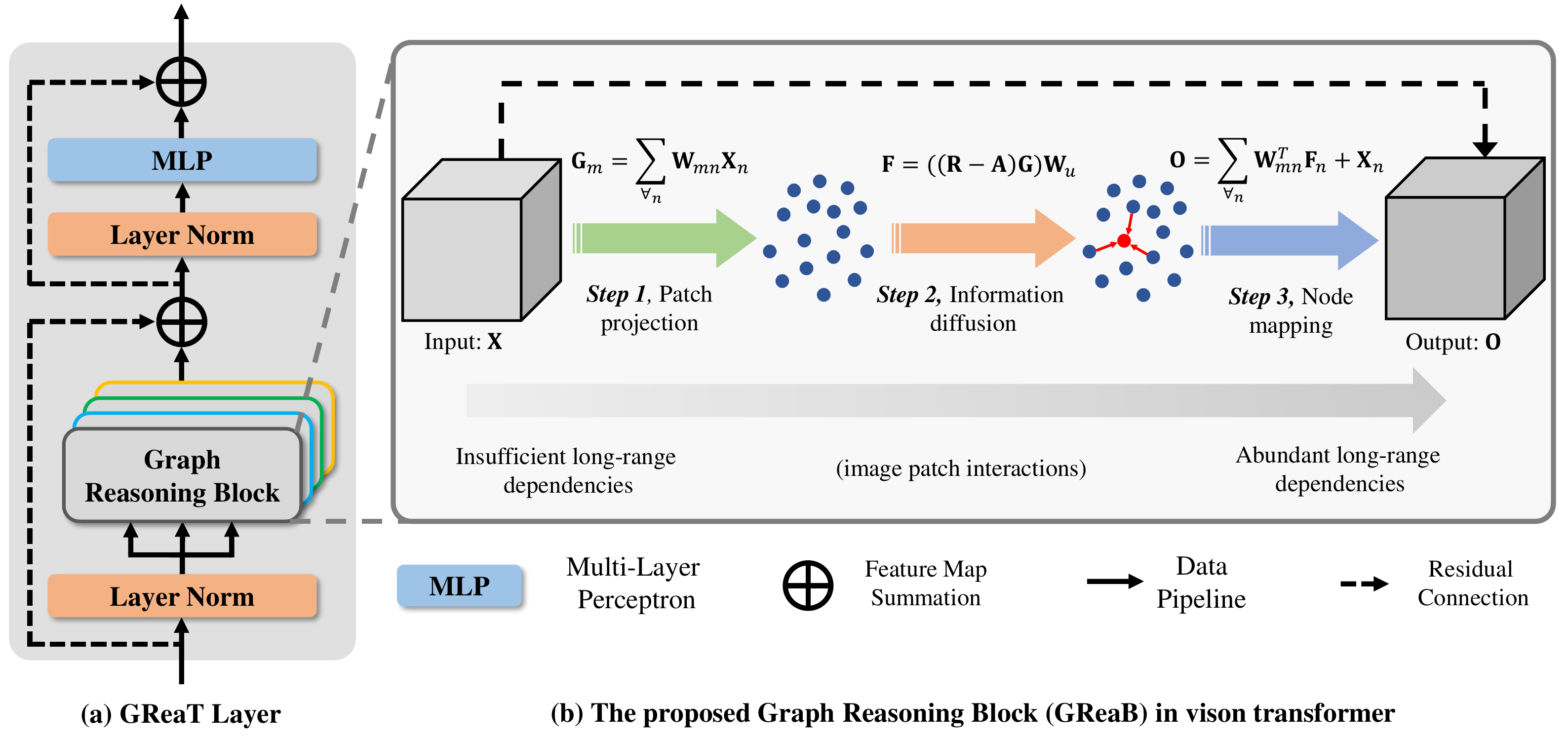}
\caption{Overall architecture of the proposed GReaT layer and the implementation details of GReaB.}
\label{fig2}
\end{figure*}

\subsection{Graph Reasoning Transformer (GReaT)}\label{sec3:2}
Although a vision transformer can eliminate the inherent locality problem in a CNN model, it potentially suffers from problems of \emph{redundant interactions of intra-class patches} and \emph{unoriented interactions of inter-class patches}. In particular, these two problems are more serious in dense prediction tasks, such as IP, because methods~\cite{zheng2021rethinking,xie2021segformer,wang2021pyramid} in this task usually adopt a relatively small patch size (\eg, $L = 4$), resulting in a large number of trivial image patches. In this work, our purpose is to alleviate these two problems and enable image patches to interact in graph space.

The model input is an RGB image $\textbf{I}$ and the output is the predicted mask $\textbf{V} \in \mathbb{R}^{H \times W \times C_{cls}}$, where each mask grid has been assigned a category label. $C_{cls}$ denotes the class size of the used dataset. Following the previous vision transformer~\cite{zheng2021rethinking,xie2021segformer,wang2021pyramid} for IP, GReaT mainly consists of a transformer encoder and a transformer decoder. For the transformer encoder, there are four ordinal Stages\footnote{In this work, we follow the common definition on ``Stage'' that features with the same spatial resolution are in the same ``Stage''~\cite{he2017mask,zheng2021rethinking,touvron2021training}.}, and features from Stage-1 to Stage-4 have the spatial resolution of $1/4$, $1/8$, $1/16$ and $1/32$ of the input, respectively. Within each Stage, as in~\cite{dosovitskiy2020image,liu2021swin,touvron2021training,wang2021pyramid}, there are several repeated transformer encoding layers. In this work, the transformer encoding layer refers to the proposed GReaT layer. As shown in Figure~\ref{fig2} (a), GReaT consists of layer norm operations, the proposed GReaB, residual connections, and an MLP. Compared to a conventional transformer encoding layer, our contribution lies in proposing a GReaB for patch interactions, \ie, replacing each attention head of the MHA operation with a GReaB. For the transformer decoder, we follow the same settings as in the previous vision transformer methods~\cite{zheng2021rethinking,xie2021segformer,liu2021swin} by using a progressive upsampling strategy or a multi-level feature aggregation strategy in our model. Implementation details of the baseline decoder are given in Section~\ref{sec4:2}.
\subsection{Graph Reasoning Block (GReaB)}\label{sec3:3}
As the core element in a GReaT layer, GReaB takes the linearly embedded image patches including the relative position encoding information as the input, and outputs a set of patches with the same scale as the input but including sufficient long-range dependencies. As illustrated in Figure~\ref{fig2} (b), GReaB contains three steps: 1) patch projection; 2) information diffusion; 3) node mapping.

\textbf{\emph{Step 1,} Patch Projection.} Patch projection aims to project image patches from the geometric space into the graph space, where each node represents an implicit visual center for a cluster of image patches. It is worth noting that each node here does not represent any specific ``instance'' or a ``category'' (\ie, the continuous visual features), but a discrete region representation. Following~\cite{chen2019graph,liang2018symbolic}, we first use the learnable patch projection weights to achieve this purpose, which can be formulated as:
\begin{equation}
\textbf{G}_m = \sum_{\forall n} \textbf{W}_{mn}\textbf{X}_n,
\label{eq4}
\end{equation}
where $\textbf{W}_{mn}\in \mathbb{R}^{1 \times L^2}$ denotes the $m$-th projection weight for the $n$-th image patch. $m$ is an index, and $m=1,2,...,M$. $M$ is the total of nodes. $\textbf{G}_m \in \mathbb{R}^{1 \times C'}$ denotes a projected node in the graph.

\textbf{\emph{Step 2,} Information Diffusion.} After obtaining $M$ nodes via patch projection, we then can establish a graph representation, where each edge reflects the relation weight between two nodes. Based on this graph, the information diffusion procedure is implemented across all nodes via a single-layer graph convolution network, which can be expressed as:
\begin{equation}
\textbf{F} = ((\textbf{R}-\textbf{A})\textbf{G})\textbf{W}_u,
\label{eq5}
\end{equation}
where $\textbf{R} \in \mathbb{R}^{M \times M}$ is an identity matrix, which is used to reduce the resistance during the model optimization stage. $\textbf{A} \in \mathbb{R}^{M \times M}$ denotes an adjacency matrix for diffusing information, which contains the relation weight between any two nodes. In our work, $\textbf{A}$ is randomly initialized and trained in an end-to-end manner along with the whole model. Following~\cite{chen2019graph,liang2018symbolic,kipf2016semi,li2018deeper}, the item $(\textbf{R}-\textbf{A})$ in information diffusion step plays a role in Laplacian smoothing. $\textbf{W}_u \in \mathbb{R}^{C' \times C'}$ denotes a trainable state update weight. Through \emph{Step 2}, the global relation information between different nodes can fully interact via this single-layer graph convolution network. Praiseworthily, since the number of graph nodes is plenarily smaller than the number of image patches, the information diffusion step has lower complexity (cf. Section~\ref{sec4:4}). In reality, we can also design the current network as a multi-layer structure (\ie, multi-layer graph convolution network). However, it will unquestionably bring significant parameter growth. Detailed trade-off analysis between the computational overheads and efficiency is given in Section~\ref{sec4:3}.

\textbf{\emph{Step 3,} Node Mapping.} After information diffusion, we map the feature representation from the graph space back into the geometry space. Considering a fact that the node mapping procedure is the reverse operation of patch projection and to reduce model parameters as much as possible, following~\cite{chen2019graph,li2018deeper}, we use the transpose of $\textbf{W}_{mn}$ for the node mapping. After a residual connection with the input, the output $\textbf{O}$ can be formulated as:
\begin{equation}
\textbf{O} = \sum_{\forall n} \textbf{W}^T_{mn}\textbf{F}_n + \textbf{X}_n,
\label{eq6}
\end{equation}
where $\textbf{F}_n \in \mathbb{R}^{1 \times C'}$ denotes the $n$-th item in $\textbf{F}$.

Compared to the multi-head attention-based patch interaction, since each node of GReaB is an intensive semantic representation for a cluster of image patches, GReaB can alleviate the problem of \emph{redundant interactions of intra-class patches}. Besides, due to the relation information among nodes in the graph-based interaction is learned, GReaB can also alleviate the problem of \emph{unoriented interactions of inter-class patches}.

\section{Experiments}
\subsection{Datasets and Evaluation Metrics}\label{sec4:1}
\myparagraph{Datasets.} In this paper, experiments are carried out on two challenging image parsing datasets, \ie, Cityscapes~\cite{mottaghi2014role} and ADE20K~\cite{zhou2017scene}.
\begin{itemize}
\item Cityscapes~\cite{cordts2016cityscapes} is a high-resolution ($1024 \times 2048$) pixel-level annotated street scene dataset by $19$ classes, which has images of $2,975$ for the \emph{training} set, $500$ for the \emph{val} set, and $1,525$ for the \emph{test} set, respectively. To make a fair comparison with other methods, we only use the finely annotated training images in our work as in~\cite{hou2020strip,huang2019ccnet}.
\item ADE20K~\cite{zhou2017scene} is one of the most challenging image parsing datasets, which contains up to $150$ classes of common scene. This dataset contains about $20$k, $2$k, and $3$k images for the \emph{training} set, \emph{val} set, and \emph{test} set, respectively.
\end{itemize}
For data augmentation on the \emph{training} set, following~\cite{zheng2021rethinking,xie2021segformer,liu2021swin,zhang2020feature}, we first use the randomly scaling  in the range of $0.5$ to $2.0$. Then, images are randomly cropped into a fixed size by $1024 \times 1024$ for Cityscapes, and by $512 \times 512$ for ADE20K. Besides, random horizontal flip and random brightness jittering are also used.

\myparagraph{Evaluation Metrics.} Following the existing methods~\cite{chen2017deeplab,zhang2019co,yuan2020object,zhang2020feature}, we use the standard mean Intersection over Union (mIoU) as the primary evaluation metric. Besides, to verify the model efficiency, the model Parameters (Params), FLOPs, and model Complexity analysis are also taken into consideration.
\subsection{Implementation Details}\label{sec4:2}

\myparagraph{Baselines.}
Three representative vision transformer IP models are chosen as baselines, \ie, SEgmentation TRansformer (SETR)~\cite{zheng2021rethinking}, SegFormer~\cite{xie2021segformer} and Swin Transformer~\cite{liu2021swin}. To assess the value of our method, we chose the stronger version of each baseline. A brief experimental setting to these three baselines is given below.
\begin{itemize}
    \item SETR~\cite{zheng2021rethinking}. A powerful encoder with $24$ layers (is named as T-Large) is set as the backbone, where the pre-trained weight is provided by~\cite{touvron2021training}. As for the transformer decoder, we choose the multi-level feature aggregation (\ie, SETR-MLA) version. Following~\cite{zhao2017pyramid,zheng2021rethinking}, the auxiliary classification loss, the synchronized batch norm in the decoder, and the multi-scale test strategy are also used.
    \item SegFormer~\cite{xie2021segformer}. The largest SegFormer-B5, where the hierarchical encoder is pre-trained on ImageNet-1K~\cite{deng2009imagenet}, is chosen as the baseline. The lightweight all MLP decoder is set as the transformer decoder and randomly initialized. Besides, the overlapped patch merging, the efficient self-attention, and the mix-FFN are used in the whole model.
    \item Swin Transformer~\cite{liu2021swin}. The powerful swin-B variant (\ie, the channel number of the hidden layer is set to $128$, and the layer number is set to $\{2,2,18,2\}$) is set as the baseline, which is pre-trained on ImageNet-22K~\cite{deng2009imagenet}. The window size is set to $7$, and the expansion layer of each MLP is set to $4$. Following its default setting, the transformer decoder is based on the hierarchical feature pyramid.
\end{itemize}

\myparagraph{Training Details.}
All models in this work including baselines are implemented on the MMSegmentation\footnote{https://github.com/open-mmlab/mmsegmentation} by using the PyTorch~\cite{paszke2019pytorch} deep learning framework on 8 NVIDIA Tesla V100 GPUs. The batch size is set to $16$ for ADE20K and $8$ for Cityscapes.

\myparagraph{Hyper-parameter Settings.} Following~\cite{zhao2017pyramid,zhang2018context,zheng2021rethinking}, the weight for auxiliary classification loss and segmentation loss is set to $0.2$ and $0.8$, respectively. In inference, the multi-scale scaling with the scaling factor of $(0.5, 0.75, 1.0, 1.25, 1.5, 1.75)$ and random horizontal flop are deployed. It's worth noting that OHEM~\cite{shrivastava2016training} and the class balance loss are not used in our model for a fair comparison.
\subsection{Ablation Study}\label{sec4:3}
The ablation study is implemented on the \emph{val} set of Cityscapes~\cite{cordts2016cityscapes}. Unless otherwise stated, the number of graph nodes $M$ is set to $16$ and the single-layer graph convolution network is adopted.
\begin{figure*}[t]
\centering
\includegraphics[width=.95\textwidth]{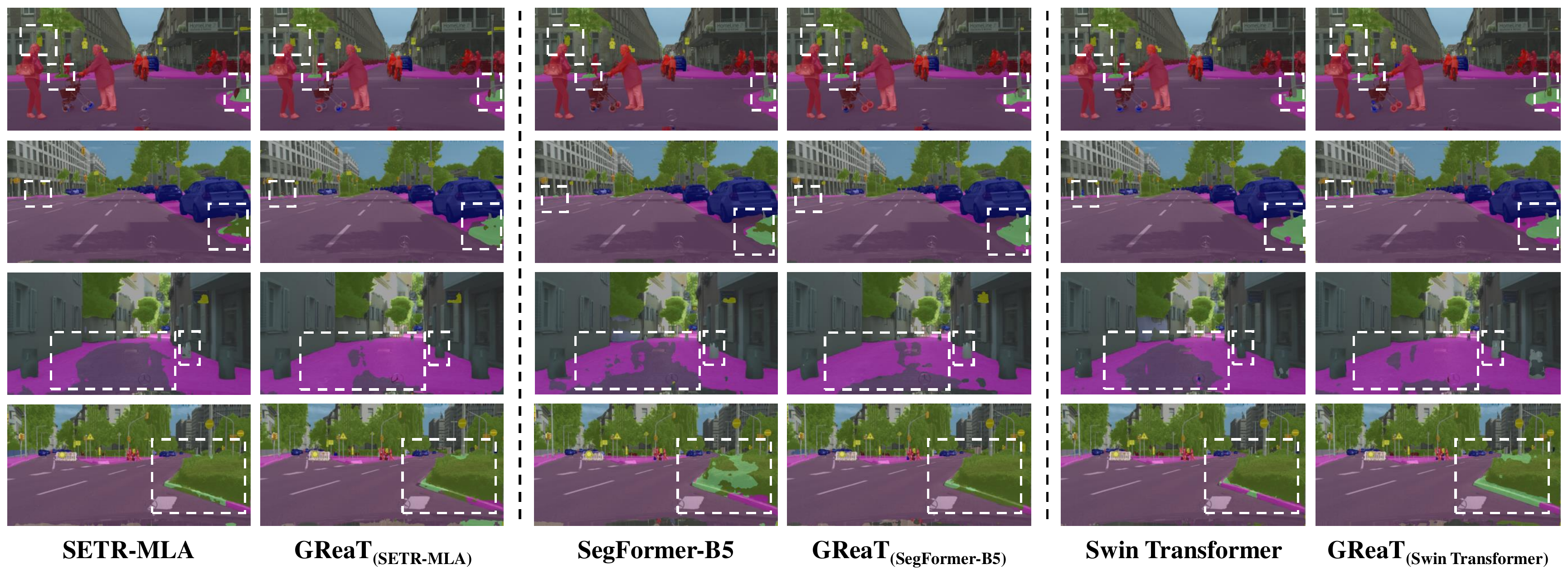}
\vspace{-3mm}
\caption{Qualitative result comparisons with baselines. The white dotted frames highlight the improved regions.}
\label{fig3}
\end{figure*}

\begin{table}[t]
\centering
\small
\renewcommand\arraystretch{1.2}
\setlength{\tabcolsep}{10pt}{
\begin{tabular}{@{}lc|cr@{}}
Method & Backbone & mIoU & \#Params. \\ 
\hline \hline
SETR-MLA & T-Large & 79.0\% & 310.6 \textbf{M} \\
GReaT$_{(\textrm{SETR-MLA})}$ & T-Large & 80.1\%$_{\color{red}{+1.1}}$ & 326.4 \textbf{M} \\
\cdashline{1-4}[1.5pt/1.5pt]
SegFormer-B5 & MiT-B5 & 83.5\% & 84.7 \textbf{M} \\
GReaT$_{(\textrm{SegFormer-B5})}$ & MiT-B5  & 84.2\%$_{\color{red}{+0.7}}$ & 94.1 \textbf{M} \\
\cdashline{1-4}[1.5pt/1.5pt]
Swin Transformer$^*$ & Swin-B & 80.2\% & 142.2 \textbf{M} \\
GReaT$_{(\textrm{Swin Transformer})}$ & Swin-B  & 81.1\%$_{\color{red}{+0.9}}$ & 150.8 \textbf{M} \\
\end{tabular}
\caption{Quantitative result comparisons with the baseline model on the \textit{val} set of Cityscapes~\cite{cordts2016cityscapes}. $*$ denotes that the results are derived based on re-implementation.}
\label{tab1}}
\end{table}
\begin{table}[t]
\centering
\small
\renewcommand\arraystretch{1.2}
\setlength{\tabcolsep}{8pt}{
\begin{tabular}{@{}l|cc|cc@{}}
Method & $M$ & Graph No. & mIoU & \#Params \\ 
\hline \hline
SETR-MLA & -- & -- & 79.0\% & 310.6 \textbf{M} \\
\hline
GReaT$_{(\textrm{SETR-MLA})}$ & 8 & 1 & 79.3\%$_{\color{red}{+0.3}}$ & 318.5 \textbf{M} \\
GReaT$_{(\textrm{SETR-MLA})}$ & 16 & 1 & 80.1\%$_{\color{red}{+1.1}}$ & 326.4 \textbf{M} \\
GReaT$_{(\textrm{SETR-MLA})}$ & 32 & 1 & 79.7\%$_{\color{red}{+0.7}}$ & 341.6 \textbf{M} \\
GReaT$_{(\textrm{SETR-MLA})}$ & 64 & 1 & 77.2\%$_{\color{blue}{-1.8}}$ & 383.0 \textbf{M} \\
\cdashline{1-5}[1.5pt/1.5pt]
GReaT$_{(\textrm{SETR-MLA})}$ & 16 & 2 & 80.4\%$_{\color{red}{+1.4}}$ & 375.2 \textbf{M} \\
GReaT$_{(\textrm{SETR-MLA})}$ & 16 & 3 & 80.5\%$_{\color{red}{+1.5}}$ & 512.6 \textbf{M} \\
\end{tabular}
\caption{The effect of the number of graph nodes and the number of graph convolutions on model performance. We show results on the \textit{val} set of Cityscapes~\cite{cordts2016cityscapes}. ``Graph No.'' denotes the number of graph layers in GReaT.}
\label{tab2}}
\vspace{-6mm}
\end{table}

\myparagraph{Effectiveness on different baselines.} We first analyze the effectiveness by implementing GReaB on different baseline models. Table~\ref{tab1} shows the performance on mIoU and Params. We can observe that GReaB can boost all the baseline performance with a slight of computational overheads. There is an average increase of $0.9\%$ mIoU on these three baselines. Specifically, with the help of GReaB, GReaT can respectively bring $1.1\%$, $0.7\%$, and $0.9\%$ mIoU improvements on SETR-MLA, SegFormer-B5, and Swin Transformer. Accordingly, the model parameters are increased by $15.8 \textbf{M} (\uparrow 5.1\%)$, $9.4 \textbf{M} (\uparrow 11.1\%)$ and $8.6 \textbf{M} (\uparrow 6.0\%)$, respectively. These results validate the effectiveness of GReaB on different baseline models and settings, and also reflect the superiority of the graph-based patch interaction in the vision transformer.

Besides, we also give a qualitative visualization analysis in Figure~\ref{fig3}. Compared to these three baselines, we can see that GReaT has more accurate prediction masks. Its superiority is embodied in some small objects (\eg, the ``indicator'', the ``roadbed'', and the distant ``grass'') and the boundary areas of some large objects (\eg, the ``sidewalk'', the ``road'', and the nearby ``meadow''). These visualization results validate that GReaB has a more productive interaction effect than the self-attention mechanism in a vision transformer.

\myparagraph{Influence of $M$.} We then analyze the influence of the number of graph nodes $M$ on the performance. SETR-MLA~\cite{zheng2021rethinking} is chosen as the unique baseline, which is the most difficult one to optimize among these baselines because of the large number of parameters. Experimental results are given in the upper part of Table~\ref{tab2}. Under the increase of $M$, we can observe that the performance shows a trend of first increasing and then decreasing on a single-layer graph convolution network. Meanwhile, the model parameters show a progressively increasing trend.
Particularly, GReaT achieves the best performance by $80.1\%$ mIoU (with 326.4 \textbf{M} Params) when $M = 16$. When $M = 64$, the performance is surprisingly lower ($\downarrow 1.8\%$ with 383.0 \textbf{M} Params) than the baseline. The reason may be that it is difficult for a graph transformer model to learn useful correlations under excessive graph nodes. Under this observation, therefore, we set $M = 16$ in the following experiments.

\myparagraph{Single-layer or multi-layer GReaB?} In the lower part of Table~\ref{tab2}, we show experimental results on the different number of graph layers (\ie, Graph No.). We can observe that as the increase of Graph No., so does the performance. The more Graph No., the greater the amount of matrix calculation overhead. Nonetheless, we summarily found that when Graph No. is greater than $1$, the performance gain vs the parameter increase is not cost-effective. Therefore, to balance the model performance and the computational overheads, we set Graph No.$ = 1$ (\ie, the single-layer GReaB) in the following experiments.
\begin{table}[t]
\centering
\small
\renewcommand\arraystretch{1.2}
\setlength{\tabcolsep}{8pt}{
\begin{tabular}{@{}l|c|cc@{}}
Method & Patch Size ($L \times L$) & mIoU & FLOPs \\ 
\hline \hline
SETR-MLA & $8 \times 8$ & 79.0\% & 2263.7 \textbf{G} \\
 \hline
GReaT$_{(\textrm{SETR-MLA})}$ & $4 \times 4$ & 79.2\%$_{\color{red}{+0.2}}$ & 2270.5 \textbf{G} \\
GReaT$_{(\textrm{SETR-MLA})}$ & $8 \times 8$ & 80.1\%$_{\color{red}{+1.1}}$ & 2261.8 \textbf{G} \\
GReaT$_{(\textrm{SETR-MLA})}$ & $16 \times 16$ & 80.0\%$_{\color{red}{+1.0}}$ & 2257.2 \textbf{G} \\
GReaT$_{(\textrm{SETR-MLA})}$ & $32 \times 32$ & 75.2\%$_{\color{blue}{-3.8}}$ & 2226.3 \textbf{G} \\
\end{tabular}
\caption{The effect of the image patch size $L \times L$ on model performance. We show results on the \textit{val} set of Cityscapes~\cite{cordts2016cityscapes}.}
\label{tab3}}
\vspace{-6mm}
\end{table}

\myparagraph{Influence of $L$.} In \emph{Step 1} of subsection~\ref{sec3:3}, $N$ image patches are projected into $M$ graph nodes. In this ablation study, we analyze the influence of image patch size $L \times L$. The baseline model is SETR-MLA~\cite{zheng2021rethinking}. Experimental results are shown in Table~\ref{tab3}. We can see that when $L$ is small (\ie, $L= 4$, $8$ and $16$), GReaT can achieve a better performance than the baseline. When $L=32$, the performance of GReaT is even worse than the baseline. The reason for this phenomenon may be that when the patch size is large, the model can not completely capture the detailed information, resulting in some critical clues being lost, which is important for the dense prediction tasks. In terms of FLOPs, it can be observed that when we set $L \geq 8$, GReaT consumes lower FLOPs than the baseline model. When we set $L = 8$, GReaT has $2261.8$ \textbf{G} FLOPs, which is $1.9$ \textbf{G} less than the baseline model. Based on these observations, we set $L = 8$ in the following experiments.
\begin{table}[t]
\centering
\small
\renewcommand\arraystretch{1.2}
\setlength{\tabcolsep}{16pt}{
\begin{tabular}{@{}l|c@{}}
Model Architecture & Space Complexity \\ 
\hline \hline
Transformer~\cite{vaswani2017attention} & $\textsl{O}(H^2W^2)$ \\
Spatial Reduction Transformer~\cite{wang2021pyramid} & $\textsl{O}(H^2W^2/r^2)$ \\
Sparse Transformer~\cite{child2019generating} & $\textsl{O}(HW\sqrt{HW})$ \\
Reformer~\cite{kitaev2020reformer} & $\textsl{O}((HW)\textrm{log}(HW))$ \\
Cross-Attention~\cite{hou2019cross} & $\textsl{O}(2(HW))$ \\
Recurrent Attention~\cite{zaremba2014recurrent} & $\textsl{O}(kHW)$ \\
Linformer~\cite{wang2020linformer} & $\textsl{O}(HW)$ \\
Performer~\cite{choromanski2020rethinking} & $\textsl{O}(HW)$ \\
LongFormer~\cite{beltagy2020longformer} & $\textsl{O}(HW)$ \\
Softmax-Free Transformer~\cite{lu2021soft} & $\textsl{O}(HW)$ \\
\cdashline{1-2}[1.5pt/1.5pt]
GReaT & $\textsl{O}(M^2)$ \\
\end{tabular}
\caption{Space complexity for various model architectures, where $r$ is the reduction rate and $k$ is the recurrent time. $M$ denotes the number of graph nodes ($M^2\ll HW$).}
\vspace{-6mm}
\label{tab4}}
\end{table}
\begin{table}[t]
\centering
\small
\renewcommand\arraystretch{1.2}
\setlength{\tabcolsep}{6pt}{
\begin{tabular}{@{}lc|ccccr@{}}
Method & Pub. Yea & Backbone & mIoU \\ 
\hline \hline
FCN (SS)~\cite{long2015fully} & CVPR'15 & ResNet-101 & 75.52\% \\
FCN (MS)~\cite{long2015fully} & CVPR'15 & ResNet-101 & 76.61\% \\
PSPNet~\cite{zhao2017pyramid} & CVPR'17 & ResNet-101 & 78.50\% \\
DeepLab-v3~\cite{chen2017rethinking} & arXiv'17 & ResNet-101 & 79.30\% \\
NonLocal~\cite{wang2018non} & CVPR'18 & ResNet-101 & 79.10\% \\
CCNet~\cite{huang2019ccnet} & CVPR'19 & ResNet-101 & 80.20\% \\
Axial-DeepLab-L~\cite{wang2020axial} & ECCV'20 & Axial-ResNet-L & 81.50\% \\
DeepLabv3+~\cite{chen2018encoder} & ECCV'18 & ResNeSt-200 & 82.70\%  \\
\cdashline{1-4}[1.5pt/1.5pt]
Swin Transformer & ICCV'21 & Swin-B & 80.20\% \\
SETR-PUP (SS)~\cite{zheng2021rethinking} & CVPR'21 & ViT-L/16 & 79.34\% \\
SETR-PUP (MS)~\cite{zheng2021rethinking} & CVPR'21 & ViT-L/16 & 82.15\% \\
Seg-B-Mask/16~\cite{strudel2021segmenter} & ICCV'21 & DeiT-B/16 & 80.60\% \\
Seg-L-Mask/16~\cite{strudel2021segmenter} & ICCV'21 & ViT-L/16 & 81.30\% \\
SegFormer~\cite{xie2021segformer} & NeurIPS'21 & MiT-B4 & 82.19\% \\
SegFormer~\cite{xie2021segformer} & NeurIPS'21 & MiT-B5 & \textbf{\color{blue}83.50\%} \\
\cdashline{1-4}[1.5pt/1.5pt]
GReaT (Ours) & ACM MM'22 & MiT-B4 & 83.02\% \\
GReaT (Ours) & ACM MM'22 & MiT-B5 & \textbf{\color{red}84.21\%} \\
\end{tabular}
\caption{Quantitative result comparisons with the state-of-the-art methods on the \textit{val} set of Cityscapes~\cite{cordts2016cityscapes}. `` Pub. Yea'' denotes the publication year with the conference name. ``SS'': Single-scale inference. ``MS'': Multi-scale inference. The \textbf{\color{red}best} and \textbf{\color{blue}second best} performance in terms of mIoU is marked with the corresponding font colors.}
\vspace{-4mm}
\label{tab5}}
\end{table}
\begin{table}[t]
\centering
\small
\renewcommand\arraystretch{1.2}
\setlength{\tabcolsep}{3pt}{
\begin{tabular}{@{}lc|ccccr@{}}
Method & Pub. Yea & Backbone & mIoU & PixAcc \\ 
\hline \hline
FCN (SS)~\cite{long2015fully} & CVPR'15 & ResNet-101 & 39.91\% & 79.52\%\\
FCN (MS)~\cite{long2015fully} & CVPR'15 & ResNet-101 & 41.40\% & 80.65\%\\
EncNet~\cite{zhang2018context} & CVPR'18 & ResNet-101 & 44.65\% & 81.69\% \\
PSPNet~\cite{zhao2017pyramid} & CVPR'17 & ResNet-269 & 44.95\% & 81.69\% \\
APCNet~\cite{he2019adaptive} & CVPR'19 & ResNet-101 & 45.38\% & - \\
OCRNet~\cite{yuan2020object} & ECCV'20 & HRNetV2-W48 & 45.45\% & - \\
SPNet~\cite{hou2020strip} & CVPR'20 & ResNet-269 & 45.60\% & 82.09\% \\
\cdashline{1-5}[1.5pt/1.5pt]
SETR-Naive (MS)~\cite{zheng2021rethinking} & CVPR'21 & T-Large & 48.80\% & 82.92\% \\
SETR-PUP (MS)~\cite{zheng2021rethinking} & CVPR'21 & T-Large & 50.09\% & 83.58\% \\
SETR-MLA (MS)~\cite{zheng2021rethinking} & CVPR'21 & T-Large & 50.28\% & 83.46\% \\
SegFormer~\cite{xie2021segformer} & NeurIPS'21 & MiT-B4 & 51.10\% & - \\
SegFormer~\cite{xie2021segformer} & NeurIPS'21 & MiT-B5 & \textbf{\color{blue}51.80\%} & - \\
\hline
GReaT (Ours) &ACM MM'22 & MiT-B4 & 51.77\% & \textbf{\color{blue}83.91\%} \\
GReaT (Ours) & ACM MM'22 & MiT-B5 & \textbf{\color{red}52.58\%} & \textbf{\color{red}84.10\%} \\
\end{tabular}
\caption{Quantitative result comparisons with the state-of-the-art methods on the \textit{val} set of ADE20K~\cite{zhou2017scene}. `` Pub. Yea'' denotes the publication year with the conference name. ``SS'': Single-scale inference. ``MS'': Multi-scale inference. The \textbf{\color{red}best} and \textbf{\color{blue}second best} performance in terms of mIoU and PixAcc is marked with the corresponding font colors.}
\vspace{-6mm}
\label{tab6}}
\end{table}
\subsection{Efficiency Analysis}\label{sec4:4}
To demonstrate the efficiency of our GReaT, we analyze the space complexity for various model architectures in Table~\ref{tab4}. For a global input token with the sequence length of $HW$, compared to the existing transformer architectures~\cite{vaswani2017attention,wang2021pyramid,child2019generating,kitaev2020reformer,hou2019cross,zaremba2014recurrent,wang2020linformer,choromanski2020rethinking,beltagy2020longformer,lu2021soft}, we can observe that GReaT has a less space complexity by only $\textsl{O}(M^2)$. For example, the classical transformer model~\cite{vaswani2017attention} has the space complexity of $\textsl{O}(H^2W^2)$, since each item of the input token participates in. Although some approaches are to shorten the length of the token via learnable sampling strategies,
some potentially critical cues may be dropped in training, such as Spatial Reduction Transformer~\cite{wang2021pyramid}, Reformer~\cite{kitaev2020reformer} and Sparse Transformer~\cite{child2019generating}. Even compared to the progressive linear transformer architectures~\cite{hou2019cross,zaremba2014recurrent,wang2020linformer,choromanski2020rethinking,beltagy2020longformer,lu2021soft}, our GReaT still has an obvious advantage in efficiency ($M^2\ll HW$). More importantly, this advantage is more pronounced when the input token has a large length.
\begin{figure*}[t]
\centering
\includegraphics[width=.95\textwidth]{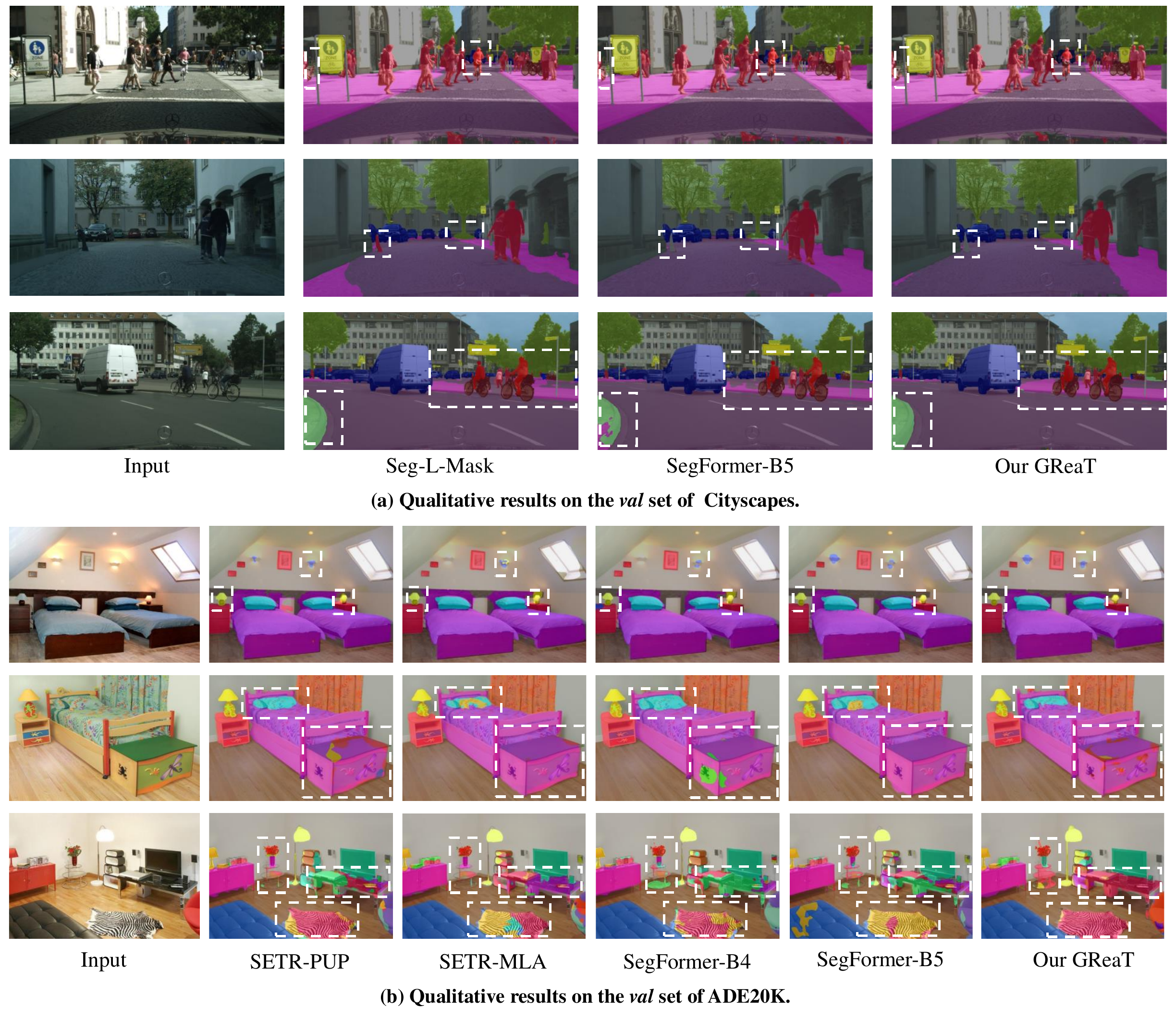}
\vspace{-4mm}
\caption{Qualitative result comparisons with the state-of-the-art methods on \emph{val} sets of Cityscapes~\cite{cordts2016cityscapes} and ADE20K~\cite{zhou2017scene}. The white dotted frames highlight the improved regions predicted by our proposed GReaT.}
\label{fig4}
\end{figure*}

\subsection{Comparisons with State-of-the-art Methods}\label{sec4:5}
In Table~\ref{tab5}, we make comparisons with the state-of-the-art methods on the \textit{val} set of Cityscapes~\cite{cordts2016cityscapes}. Our proposed GReaT achieves the competitive mIoU of $83.02\%$ with MiT-B4 as the backbone, which surpasses the baseline SegFormer~\cite{xie2021segformer} with MiT-B4 by $0.93\%$ mIoU. When we adopted MiT-B5 as the backbone, GReaT can achieve the mIoU of $84.21\%$, which demonstrates that our model brings consistent improvements on a stronger backbone. Result comparisons with state-of-the-art methods on the \textit{val} set of ADE20K are shown in Table~\ref{tab6}. We can see that under the help of GReaT, the model performance consistently improves as well. We finally achieve $51.77\%$ and $52.58\%$ mIoU, and $83.91\%$ and $84.10\%$ PixAcc on ADE20K.

Besides, we also show some qualitative visualization comparisons with the state-of-the-art methods on \emph{val} sets of Cityscapes~\cite{cordts2016cityscapes} and ADE20K~\cite{zhou2017scene} in Figure~\ref{fig4}. The progressive Seg-L-Mask~\cite{strudel2021segmenter}, SegFormer\cite{xie2021segformer} and SETR~\cite{zheng2021rethinking} are used for the result comparison. We observe that our proposed GReaT achieves better segmentation mask predictions on some small objects (\eg, the ``person'', the ``lamp'' and the ``flower basket''), large objects (\eg, the ``pillow'', the ``footstool'' and the ``TV bench'') and object boundaries (\eg, the ``sidewalk'' and the ``carpet'') at the same time.

\section{Conclusion}
In this work, we address the problems of \emph{redundant interactions of intra-class patches} and \emph{unoriented interactions of inter-class patches} in the existing vision transformer. We propose a GReaT which enables image patches to interact in the graph space following a relation reasoning pattern. GReaT has higher interaction efficiency and a more purposeful interaction pattern than the conventional transformer. Experimental results on two challenging IP datasets validate that GReaT can bring consistent performance gains with a slight computational overhead. GReaT being a general model for vision transformer, we plan to apply it to some other computer vision tasks, \eg, object detection, object localization, person re-identification, and image generation in the future. 

\section*{Acknowledgments}
The authors would like to thank all the anonymous reviewers for their constructive comments and suggestions. This work was partially supported by the National Key Research and Development Program of China under Grant 2018AAA0102002, the National Science Fund for Distinguished Young Scholars under Grant 61925204, ACCESS – AI Chip Center for Emerging Smart Systems, sponsored by InnoHK funding, Hong Kong SAR. 
\bibliographystyle{plain}
\bibliography{sample-base}

\begin{thebibliography}{10}

\bibitem{badrinarayanan2017segnet}
Vijay Badrinarayanan, Alex Kendall, and Roberto Cipolla.
\newblock Segnet: A deep convolutional encoder-decoder architecture for image
  segmentation.
\newblock {\em IEEE Transactions on Pattern Analysis and Machine Intelligence},
  39(12):2481--2495, 2017.

\bibitem{beltagy2020longformer}
Iz~Beltagy, Matthew~E Peters, and Arman Cohan.
\newblock Longformer: The long-document transformer.
\newblock In {\em arXiv}, 2020.

\bibitem{bertasius2017convolutional}
Gedas Bertasius, Lorenzo Torresani, Stella~X Yu, and Jianbo Shi.
\newblock Convolutional random walk networks for semantic image segmentation.
\newblock In {\em Proceedings of the IEEE Conference on Computer Vision and
  Pattern Recognition (CVPR)}, 2017.

\bibitem{buades2005non}
Antoni Buades, Bartomeu Coll, and J-M Morel.
\newblock A non-local algorithm for image denoising.
\newblock In {\em Proceedings of the IEEE Conference on Computer Vision and
  Pattern Recognition (CVPR)}, 2005.

\bibitem{carion2020end}
Nicolas Carion, Francisco Massa, Gabriel Synnaeve, Nicolas Usunier, Alexander
  Kirillov, and Sergey Zagoruyko.
\newblock End-to-end object detection with transformers.
\newblock In {\em European Conference on Computer Vision (ECCV)}, 2020.

\bibitem{chen2021transunet}
Jieneng Chen, Yongyi Lu, Qihang Yu, Xiangde Luo, Ehsan Adeli, Yan Wang, Le~Lu,
  Alan~L Yuille, and Yuyin Zhou.
\newblock Transunet: Transformers make strong encoders for medical image
  segmentation.
\newblock In {\em arXiv}, 2021.

\bibitem{chen2017deeplab}
Liang-Chieh Chen, George Papandreou, Iasonas Kokkinos, Kevin Murphy, and Alan~L
  Yuille.
\newblock Deeplab: Semantic image segmentation with deep convolutional nets,
  atrous convolution, and fully connected crfs.
\newblock {\em IEEE Transactions on Pattern Analysis and Machine Intelligence},
  40(4):834--848, 2017.

\bibitem{chen2017rethinking}
Liang-Chieh Chen, George Papandreou, Florian Schroff, and Hartwig Adam.
\newblock Rethinking atrous convolution for semantic image segmentation.
\newblock In {\em arXiv}, 2017.

\bibitem{chen2018encoder}
Liang-Chieh Chen, Yukun Zhu, George Papandreou, Florian Schroff, and Hartwig
  Adam.
\newblock Encoder-decoder with atrous separable convolution for semantic image
  segmentation.
\newblock In {\em European Conference on Computer Vision (ECCV)}, 2018.

\bibitem{chen2019graph}
Yunpeng Chen, Marcus Rohrbach, Zhicheng Yan, Yan Shuicheng, Jiashi Feng, and
  Yannis Kalantidis.
\newblock Graph-based global reasoning networks.
\newblock In {\em Proceedings of the IEEE Conference on Computer Vision and
  Pattern Recognition (CVPR)}, 2019.

\bibitem{child2019generating}
Rewon Child, Scott Gray, Alec Radford, and Ilya Sutskever.
\newblock Generating long sequences with sparse transformers.
\newblock In {\em arXiv}, 2019.

\bibitem{choe2020attention}
Junsuk Choe, Seungho Lee, and Hyunjung Shim.
\newblock Attention-based dropout layer for weakly supervised single object
  localization and semantic segmentation.
\newblock {\em IEEE Transactions on Pattern Analysis and Machine Intelligence},
  43(12):4256--4271, 2020.

\bibitem{choromanski2020rethinking}
Krzysztof Choromanski, Valerii Likhosherstov, David Dohan, Xingyou Song,
  Andreea Gane, Tamas Sarlos, Peter Hawkins, Jared Davis, Afroz Mohiuddin,
  Lukasz Kaiser, et~al.
\newblock Rethinking attention with performers.
\newblock In {\em International Conference on Learning Representations (ICLR)},
  2021.

\bibitem{cordts2016cityscapes}
Marius Cordts, Mohamed Omran, Sebastian Ramos, Timo Rehfeld, Markus Enzweiler,
  Rodrigo Benenson, Uwe Franke, Stefan Roth, and Bernt Schiele.
\newblock The cityscapes dataset for semantic urban scene understanding.
\newblock In {\em Proceedings of the IEEE Conference on Computer Vision and
  Pattern Recognition (CVPR)}, 2016.

\bibitem{deng2021spatially}
Han Deng, Chu Han, Hongmin Cai, Guoqiang Han, and Shengfeng He.
\newblock Spatially-invariant style-codes controlled makeup transfer.
\newblock In {\em Proceedings of the IEEE Conference on Computer Vision and
  Pattern Recognition (CVPR)}, 2021.

\bibitem{deng2014large}
Jia Deng, Nan Ding, Yangqing Jia, Andrea Frome, Kevin Murphy, Samy Bengio, Yuan
  Li, Hartmut Neven, and Hartwig Adam.
\newblock Large-scale object classification using label relation graphs.
\newblock In {\em European Conference on Computer Vision (ECCV)}, 2014.

\bibitem{deng2009imagenet}
Jia Deng, Wei Dong, Richard Socher, Li-Jia Li, Kai Li, and Li~Fei-Fei.
\newblock Imagenet: A large-scale hierarchical image database.
\newblock In {\em Proceedings of the IEEE Conference on Computer Vision and
  Pattern Recognition (CVPR)}, 2009.

\bibitem{devlin2018bert}
Jacob Devlin, Ming-Wei Chang, Kenton Lee, and Kristina Toutanova.
\newblock Bert: Pre-training of deep bidirectional transformers for language
  understanding.
\newblock In {\em arXiv}, 2018.

\bibitem{dosovitskiy2020image}
Alexey Dosovitskiy, Lucas Beyer, Alexander Kolesnikov, Dirk Weissenborn,
  Xiaohua Zhai, Thomas Unterthiner, Mostafa Dehghani, Matthias Minderer, Georg
  Heigold, Sylvain Gelly, et~al.
\newblock An image is worth 16x16 words: Transformers for image recognition at
  scale.
\newblock In {\em International Conference on Learning Representations (ICLR)},
  2020.

\bibitem{fu2019dual}
Jun Fu, Jing Liu, Haijie Tian, Yong Li, Yongjun Bao, Zhiwei Fang, and Hanqing
  Lu.
\newblock Dual attention network for scene segmentation.
\newblock In {\em Proceedings of the IEEE/CVF International Conference on
  Computer Vision (ICCV)}, 2019.

\bibitem{ghiasi2018dropblock}
Golnaz Ghiasi, Tsung-Yi Lin, and Quoc~V Le.
\newblock Dropblock: A regularization method for convolutional networks.
\newblock In {\em Advances in Neural Information Processing Systems (NeurIPS)},
  2018.

\bibitem{gulati2020conformer}
Anmol Gulati, James Qin, Chung-Cheng Chiu, Niki Parmar, Yu~Zhang, Jiahui Yu,
  Wei Han, Shibo Wang, Zhengdong Zhang, Yonghui Wu, et~al.
\newblock Conformer: Convolution-augmented transformer for speech recognition.
\newblock In {\em arXiv}, 2020.

\bibitem{he2019adaptive}
Junjun He, Zhongying Deng, Lei Zhou, Yali Wang, and Yu~Qiao.
\newblock Adaptive pyramid context network for semantic segmentation.
\newblock In {\em Proceedings of the IEEE Conference on Computer Vision and
  Pattern Recognition (CVPR)}, 2019.

\bibitem{he2017mask}
Kaiming He, Georgia Gkioxari, Piotr Doll{\'a}r, and Ross Girshick.
\newblock Mask r-cnn.
\newblock In {\em Proceedings of the IEEE/CVF International Conference on
  Computer Vision (ICCV)}, 2017.

\bibitem{he2016deep}
Kaiming He, Xiangyu Zhang, Shaoqing Ren, and Jian Sun.
\newblock Deep residual learning for image recognition.
\newblock In {\em Proceedings of the IEEE Conference on Computer Vision and
  Pattern Recognition (CVPR)}, 2016.

\bibitem{hou2020strip}
Qibin Hou, Li~Zhang, Ming-Ming Cheng, and Jiashi Feng.
\newblock Strip pooling: Rethinking spatial pooling for scene parsing.
\newblock In {\em Proceedings of the IEEE Conference on Computer Vision and
  Pattern Recognition (CVPR)}, 2020.

\bibitem{hou2019cross}
Ruibing Hou, Hong Chang, Bingpeng Ma, Shiguang Shan, and Xilin Chen.
\newblock Cross attention network for few-shot classification.
\newblock In {\em Advances in Neural Information Processing Systems (NeurIPS)},
  2019.

\bibitem{huang2019ccnet}
Zilong Huang, Xinggang Wang, Lichao Huang, Chang Huang, Yunchao Wei, and Wenyu
  Liu.
\newblock Ccnet: Criss-cross attention for semantic segmentation.
\newblock In {\em Proceedings of the IEEE Conference on Computer Vision and
  Pattern Recognition (CVPR)}, 2019.

\bibitem{jain2021representing}
Paras Jain, Zhanghao Wu, Matthew Wright, Azalia Mirhoseini, Joseph~E Gonzalez,
  and Ion Stoica.
\newblock Representing long-range context for graph neural networks with global
  attention.
\newblock In {\em Advances in Neural Information Processing Systems (NeurIPS)},
  2021.

\bibitem{kipf2016semi}
Thomas~N Kipf and Max Welling.
\newblock Semi-supervised classification with graph convolutional networks.
\newblock In {\em International Conference on Learning Representations (ICLR)},
  2016.

\bibitem{kitaev2020reformer}
Nikita Kitaev, {\L}ukasz Kaiser, and Anselm Levskaya.
\newblock Reformer: The efficient transformer.
\newblock In {\em International Conference on Learning Representations (ICLR)},
  2020.

\bibitem{lecun2015deep}
Yann LeCun, Yoshua Bengio, and Geoffrey Hinton.
\newblock Deep learning.
\newblock {\em Nature}, 521(7553):436--444, 2015.

\bibitem{li2018deeper}
Qimai Li, Zhichao Han, and Xiao-Ming Wu.
\newblock Deeper insights into graph convolutional networks for semi-supervised
  learning.
\newblock In {\em Association for the Advancement of Artificial Intelligence
  (AAAI)}, 2018.

\bibitem{li2021superpixel}
Shuailin Li, Zhitong Gao, and Xuming He.
\newblock Superpixel-guided iterative learning from noisy labels for medical
  image segmentation.
\newblock In {\em International Conference on Medical Image Computing and
  Computer-Assisted Intervention (MICCAI)}, 2021.

\bibitem{li2018beyond}
Yin Li and Abhinav Gupta.
\newblock Beyond grids: Learning graph representations for visual recognition.
\newblock In {\em Advances in Neural Information Processing Systems (NeurIPS)},
  2018.

\bibitem{liang2018symbolic}
Xiaodan Liang, Zhiting Hu, Hao Zhang, Liang Lin, and Eric~P Xing.
\newblock Symbolic graph reasoning meets convolutions.
\newblock In {\em Advances in Neural Information Processing Systems (NeurIPS)},
  2018.

\bibitem{liang2022not}
Youwei Liang, Chongjian Ge, Zhan Tong, Yibing Song, Jue Wang, and Pengtao Xie.
\newblock Not all patches are what you need: Expediting vision transformers via
  token reorganizations.
\newblock In {\em International Conference on Learning Representations (ICLR)},
  2022.

\bibitem{lin2017refinenet}
Guosheng Lin, Anton Milan, Chunhua Shen, and Ian Reid.
\newblock Refinenet: Multi-path refinement networks for high-resolution
  semantic segmentation.
\newblock In {\em Proceedings of the IEEE Conference on Computer Vision and
  Pattern Recognition (CVPR)}, 2017.

\bibitem{lin2020graphonomy}
Liang Lin, Yiming Gao, Ke~Gong, Meng Wang, and Xiaodan Liang.
\newblock Graphonomy: Universal image parsing via graph reasoning and transfer.
\newblock {\em IEEE Transactions on Pattern Analysis and Machine Intelligence},
  2020.

\bibitem{liu2021swin}
Ze~Liu, Yutong Lin, Yue Cao, Han Hu, Yixuan Wei, Zheng Zhang, Stephen Lin, and
  Baining Guo.
\newblock Swin transformer: Hierarchical vision transformer using shifted
  windows.
\newblock In {\em Proceedings of the IEEE Conference on Computer Vision and
  Pattern Recognition (CVPR)}, 2021.

\bibitem{long2015fully}
Jonathan Long, Evan Shelhamer, and Trevor Darrell.
\newblock Fully convolutional networks for semantic segmentation.
\newblock In {\em Proceedings of the IEEE Conference on Computer Vision and
  Pattern Recognition (CVPR)}, 2015.

\bibitem{lu2021soft}
Jiachen Lu, Jinghan Yao, Junge Zhang, Xiatian Zhu, Hang Xu, Weiguo Gao,
  Chunjing Xu, Tao Xiang, and Li~Zhang.
\newblock Soft: Softmax-free transformer with linear complexity.
\newblock In {\em Advances in Neural Information Processing Systems (NeurIPS)},
  2021.

\bibitem{meinhardt2021trackformer}
Tim Meinhardt, Alexander Kirillov, Laura Leal-Taixe, and Christoph
  Feichtenhofer.
\newblock Trackformer: Multi-object tracking with transformers.
\newblock In {\em arXiv}, 2021.

\bibitem{mottaghi2014role}
Roozbeh Mottaghi, Xianjie Chen, Xiaobai Liu, Nam-Gyu Cho, Seong-Whan Lee, Sanja
  Fidler, Raquel Urtasun, and Alan Yuille.
\newblock The role of context for object detection and semantic segmentation in
  the wild.
\newblock In {\em Proceedings of the IEEE Conference on Computer Vision and
  Pattern Recognition (CVPR)}, 2014.

\bibitem{paszke2019pytorch}
Adam Paszke, Sam Gross, Francisco Massa, Adam Lerer, James Bradbury, Gregory
  Chanan, Trevor Killeen, Zeming Lin, Natalia Gimelshein, Luca Antiga, et~al.
\newblock Pytorch: An imperative style, high-performance deep learning library.
\newblock In {\em Advances in Neural Information Processing Systems (NeurIPS)},
  2019.

\bibitem{peng2017large}
Chao Peng, Xiangyu Zhang, Gang Yu, Guiming Luo, and Jian Sun.
\newblock Large kernel matters--improve semantic segmentation by global
  convolutional network.
\newblock In {\em Proceedings of the IEEE Conference on Computer Vision and
  Pattern Recognition (CVPR)}, 2017.

\bibitem{plath2009multi}
Nils Plath, Marc Toussaint, and Shinichi Nakajima.
\newblock Multi-class image segmentation using conditional random fields and
  global classification.
\newblock In {\em International Conference on Machine Learning (ICML)}, 2009.

\bibitem{prakash2021multi}
Aditya Prakash, Kashyap Chitta, and Andreas Geiger.
\newblock Multi-modal fusion transformer for end-to-end autonomous driving.
\newblock In {\em Proceedings of the IEEE Conference on Computer Vision and
  Pattern Recognition (CVPR)}, 2021.

\bibitem{redmon2017yolo9000}
Joseph Redmon and Ali Farhadi.
\newblock Yolo9000: better, faster, stronger.
\newblock In {\em Proceedings of the IEEE Conference on Computer Vision and
  Pattern Recognition (CVPR)}, 2017.

\bibitem{ren2015faster}
Shaoqing Ren, Kaiming He, Ross Girshick, and Jian Sun.
\newblock Faster r-cnn: Towards real-time object detection with region proposal
  networks.
\newblock In {\em Advances in Neural Information Processing Systems (NeurIPS)},
  2015.

\bibitem{rong2020self}
Yu~Rong, Yatao Bian, Tingyang Xu, Weiyang Xie, Ying Wei, Wenbing Huang, and
  Junzhou Huang.
\newblock Self-supervised graph transformer on large-scale molecular data.
\newblock In {\em Advances in Neural Information Processing Systems (NeurIPS)},
  2020.

\bibitem{ronneberger2015u}
Olaf Ronneberger, Philipp Fischer, and Thomas Brox.
\newblock U-net: Convolutional networks for biomedical image segmentation.
\newblock In {\em International Conference on Medical Image Computing and
  Computer Assisted Intervention (MICCAI)}, 2015.

\bibitem{sabour2017dynamic}
Sara Sabour, Nicholas Frosst, and Geoffrey~E Hinton.
\newblock Dynamic routing between capsules.
\newblock In {\em Advances in Neural Information Processing Systems (NeurIPS)},
  2017.

\bibitem{shrivastava2016training}
Abhinav Shrivastava, Abhinav Gupta, and Ross Girshick.
\newblock Training region-based object detectors with online hard example
  mining.
\newblock In {\em Proceedings of the IEEE Conference on Computer Vision and
  Pattern Recognition (CVPR)}, 2016.

\bibitem{strudel2021segmenter}
Robin Strudel, Ricardo Garcia, Ivan Laptev, and Cordelia Schmid.
\newblock Segmenter: Transformer for semantic segmentation.
\newblock In {\em Proceedings of the IEEE/CVF International Conference on
  Computer Vision (ICCV)}, 2021.

\bibitem{tan2019efficientnet}
Mingxing Tan and Quoc Le.
\newblock Efficientnet: Rethinking model scaling for convolutional neural
  networks.
\newblock In {\em International Conference on Machine Learning (ICML)}, 2019.

\bibitem{TangLPT20}
Hao Tang, Zechao Li, Zhimao Peng, and Jinhui Tang.
\newblock Blockmix: Meta regularization and self-calibrated inference for
  metric-based meta-learning.
\newblock In {\em ACM international conference on Multimedia (MM)}, 2020.

\bibitem{TANG2022108792}
Hao Tang, Chengcheng Yuan, Zechao Li, and Jinhui Tang.
\newblock Learning attention-guided pyramidal features for few-shot
  fine-grained recognition.
\newblock {\em Pattern Recognition}, 130:108792, 2022.

\bibitem{tian2019decoders}
Zhi Tian, Tong He, Chunhua Shen, and Youliang Yan.
\newblock Decoders matter for semantic segmentation: Data-dependent decoding
  enables flexible feature aggregation.
\newblock In {\em Proceedings of the IEEE Conference on Computer Vision and
  Pattern Recognition (CVPR)}, 2019.

\bibitem{tompson2015efficient}
Jonathan Tompson, Ross Goroshin, Arjun Jain, Yann LeCun, and Christoph Bregler.
\newblock Efficient object localization using convolutional networks.
\newblock In {\em Proceedings of the IEEE Conference on Computer Vision and
  Pattern Recognition (CVPR)}, 2015.

\bibitem{touvron2021training}
Hugo Touvron, Matthieu Cord, Matthijs Douze, Francisco Massa, Alexandre
  Sablayrolles, and Herv{\'e} J{\'e}gou.
\newblock Training data-efficient image transformers \& distillation through
  attention.
\newblock In {\em International Conference on Machine Learning (ICML)}, 2021.

\bibitem{vaswani2017attention}
Ashish Vaswani, Noam Shazeer, Niki Parmar, Jakob Uszkoreit, Llion Jones,
  Aidan~N Gomez, {\L}ukasz Kaiser, and Illia Polosukhin.
\newblock Attention is all you need.
\newblock In {\em Advances in Neural Information Processing Systems (NeurIPS)},
  2017.

\bibitem{Wang_2022_IJCAI}
Di~Wang, Jinyuan Liu, Xin Fan, and Risheng Liu.
\newblock Unsupervised misaligned infrared and visible image fusion via
  cross-modality image generation and registration.
\newblock {\em CoRR}, abs/2205.11876, 2022.

\bibitem{wang2020axial}
Huiyu Wang, Yukun Zhu, Bradley Green, Hartwig Adam, Alan Yuille, and
  Liang-Chieh Chen.
\newblock Axial-deeplab: Stand-alone axial-attention for panoptic segmentation.
\newblock In {\em European Conference on Computer Vision (ECCV)}, 2020.

\bibitem{wang2020deep}
Jingdong Wang, Ke~Sun, Tianheng Cheng, Borui Jiang, Chaorui Deng, Yang Zhao,
  Dong Liu, Yadong Mu, Mingkui Tan, Xinggang Wang, et~al.
\newblock Deep high-resolution representation learning for visual recognition.
\newblock {\em IEEE Transactions on Pattern Analysis and Machine Intelligence},
  43(10):3349--3364, 2020.

\bibitem{wang2020linformer}
Sinong Wang, Belinda~Z Li, Madian Khabsa, Han Fang, and Hao Ma.
\newblock Linformer: Self-attention with linear complexity.
\newblock In {\em arXiv}, 2020.

\bibitem{wang2020visual}
Tan Wang, Jianqiang Huang, Hanwang Zhang, and Qianru Sun.
\newblock Visual commonsense r-cnn.
\newblock In {\em Proceedings of the IEEE Conference on Computer Vision and
  Pattern Recognition (CVPR)}, 2020.

\bibitem{wang2021pyramid}
Wenhai Wang, Enze Xie, Xiang Li, Deng-Ping Fan, Kaitao Song, Ding Liang, Tong
  Lu, Ping Luo, and Ling Shao.
\newblock Pyramid vision transformer: A versatile backbone for dense prediction
  without convolutions.
\newblock In {\em Proceedings of the IEEE Conference on Computer Vision and
  Pattern Recognition (CVPR)}, 2021.

\bibitem{wang2018non}
Xiaolong Wang, Ross Girshick, Abhinav Gupta, and Kaiming He.
\newblock Non-local neural networks.
\newblock In {\em Proceedings of the IEEE Conference on Computer Vision and
  Pattern Recognition (CVPR)}, 2018.

\bibitem{wang2018videos}
Xiaolong Wang and Abhinav Gupta.
\newblock Videos as space-time region graphs.
\newblock In {\em European Conference on Computer Vision (ECCV)}, 2018.

\bibitem{wang2021not}
Yulin Wang, Rui Huang, Shiji Song, Zeyi Huang, and Gao Huang.
\newblock Not all images are worth 16x16 words: Dynamic transformers for
  efficient image recognition.
\newblock In {\em Advances in Neural Information Processing Systems (NeurIPS)},
  2021.

\bibitem{wang2021end}
Yuqing Wang, Zhaoliang Xu, Xinlong Wang, Chunhua Shen, Baoshan Cheng, Hao Shen,
  and Huaxia Xia.
\newblock End-to-end video instance segmentation with transformers.
\newblock In {\em Proceedings of the IEEE Conference on Computer Vision and
  Pattern Recognition (CVPR)}, 2021.

\bibitem{xie2021segformer}
Enze Xie, Wenhai Wang, Zhiding Yu, Anima Anandkumar, Jose~M Alvarez, and Ping
  Luo.
\newblock Segformer: Simple and efficient design for semantic segmentation with
  transformers.
\newblock In {\em Advances in Neural Information Processing Systems (NeurIPS)},
  2021.

\bibitem{xie2017aggregated}
Saining Xie, Ross Girshick, Piotr Doll{\'a}r, Zhuowen Tu, and Kaiming He.
\newblock Aggregated residual transformations for deep neural networks.
\newblock In {\em Proceedings of the IEEE Conference on Computer Vision and
  Pattern Recognition (CVPR)}, 2017.

\bibitem{yan2018participation}
Rui Yan, Jinhui Tang, Xiangbo Shu, Zechao Li, and Qi~Tian.
\newblock Participation-contributed temporal dynamic model for group activity
  recognition.
\newblock In {\em ACM international conference on Multimedia (MM)}, 2018.

\bibitem{yan2020higcin}
Rui Yan, Lingxi Xie, Jinhui Tang, Xiangbo Shu, and Qi~Tian.
\newblock Higcin: Hierarchical graph-based cross inference network for group
  activity recognition.
\newblock {\em IEEE Transactions on Pattern Analysis and Machine Intelligence},
  2020.

\bibitem{yan2020social}
Rui Yan, Lingxi Xie, Jinhui Tang, Xiangbo Shu, and Qi~Tian.
\newblock Social adaptive module for weakly-supervised group activity
  recognition.
\newblock In {\em European Conference on Computer Vision (ECCV)}, 2020.

\bibitem{yang2021focal}
Jianwei Yang, Chunyuan Li, Pengchuan Zhang, Xiyang Dai, Bin Xiao, Lu~Yuan, and
  Jianfeng Gao.
\newblock Focal self-attention for local-global interactions in vision
  transformers.
\newblock In {\em Advances in Neural Information Processing Systems (NeurIPS)},
  2021.

\bibitem{yu2015multi}
Fisher Yu and Vladlen Koltun.
\newblock Multi-scale context aggregation by dilated convolutions.
\newblock In {\em International Conference on Learning Representations (ICLR)},
  2015.

\bibitem{yuan2020object}
Yuhui Yuan, Xilin Chen, and Jingdong Wang.
\newblock Object-contextual representations for semantic segmentation.
\newblock In {\em European Conference on Computer Vision (ECCV)}, 2020.

\bibitem{zaremba2014recurrent}
Wojciech Zaremba, Ilya Sutskever, and Oriol Vinyals.
\newblock Recurrent neural network regularization.
\newblock In {\em arXi}, 2014.

\bibitem{zhang2020causal}
Dong Zhang, Hanwang Zhang, Jinhui Tang, Xian-Sheng Hua, and Qianru Sun.
\newblock Causal intervention for weakly-supervised semantic segmentation.
\newblock In {\em Advances in Neural Information Processing Systems (NeurIPS)},
  2020.

\bibitem{zhang2021self}
Dong Zhang, Hanwang Zhang, Jinhui Tang, Xian-Sheng Hua, and Qianru Sun.
\newblock Self-regulation for semantic segmentation.
\newblock In {\em Proceedings of the IEEE/CVF International Conference on
  Computer Vision (ICCV)}, 2021.

\bibitem{zhang2020feature}
Dong Zhang, Hanwang Zhang, Jinhui Tang, Meng Wang, Xiansheng Hua, and Qianru
  Sun.
\newblock Feature pyramid transformer.
\newblock In {\em European Conference on Computer Vision (ECCV)}, 2020.

\bibitem{zhang2018context}
Hang Zhang, Kristin Dana, Jianping Shi, Zhongyue Zhang, Xiaogang Wang, Ambrish
  Tyagi, and Amit Agrawal.
\newblock Context encoding for semantic segmentation.
\newblock In {\em Proceedings of the IEEE Conference on Computer Vision and
  Pattern Recognition (CVPR)}, 2018.

\bibitem{zhang2019co}
Hang Zhang, Han Zhang, Chenguang Wang, and Junyuan Xie.
\newblock Co-occurrent features in semantic segmentation.
\newblock In {\em Proceedings of the IEEE Conference on Computer Vision and
  Pattern Recognition (CVPR)}, 2019.

\bibitem{zhang2020accelerating}
Minjia Zhang and Yuxiong He.
\newblock Accelerating training of transformer-based language models with
  progressive layer dropping.
\newblock In {\em Advances in Neural Information Processing Systems (NeurIPS)},
  2020.

\bibitem{zhao2017open}
Hang Zhao, Xavier Puig, Bolei Zhou, Sanja Fidler, and Antonio Torralba.
\newblock Open vocabulary scene parsing.
\newblock In {\em Proceedings of the IEEE/CVF International Conference on
  Computer Vision (ICCV)}, 2017.

\bibitem{zhao2017pyramid}
Hengshuang Zhao, Jianping Shi, Xiaojuan Qi, Xiaogang Wang, and Jiaya Jia.
\newblock Pyramid scene parsing network.
\newblock In {\em Proceedings of the IEEE Conference on Computer Vision and
  Pattern Recognition (CVPR)}, 2017.

\bibitem{zheng2021rethinking}
Sixiao Zheng, Jiachen Lu, Hengshuang Zhao, Xiatian Zhu, Zekun Luo, Yabiao Wang,
  Yanwei Fu, Jianfeng Feng, Tao Xiang, Philip~HS Torr, et~al.
\newblock Rethinking semantic segmentation from a sequence-to-sequence
  perspective with transformers.
\newblock In {\em Proceedings of the IEEE Conference on Computer Vision and
  Pattern Recognition (CVPR)}, 2021.

\bibitem{zhou2017scene}
Bolei Zhou, Hang Zhao, Xavier Puig, Sanja Fidler, Adela Barriuso, and Antonio
  Torralba.
\newblock Scene parsing through ade20k dataset.
\newblock In {\em Proceedings of the IEEE Conference on Computer Vision and
  Pattern Recognition (CVPR)}, 2017.

\bibitem{zhou2021nnformer}
Hong-Yu Zhou, Jiansen Guo, Yinghao Zhang, Lequan Yu, Liansheng Wang, and Yizhou
  Yu.
\newblock nnformer: Interleaved transformer for volumetric segmentation.
\newblock In {\em arXiv}, 2021.

\end{thebibliography}
\end{document}